\documentclass[10pt,twocolumn,letterpaper]{article}

\usepackage[pagenumbers]{cvpr}
\usepackage{cuted} %

\usepackage{currfile} %

\usepackage{caption} %
 \usepackage[protrusion=true,expansion=true]{microtype}
\makeatletter
\renewcommand\paragraph{\@startsection{paragraph}{4}{\z@}{0.5ex \@plus 0.2ex \@minus 0.1ex}{-0.8em}{\normalfont\normalsize\bfseries}}
\makeatother

\definecolor{cvprblue}{rgb}{0.21,0.49,0.74}
\usepackage[pagebackref,breaklinks,colorlinks,allcolors=cvprblue]{hyperref}
\usepackage{booktabs}
\usepackage{multirow}
\usepackage{graphicx}
\usepackage{amsmath}
\usepackage{amssymb}
\usepackage{xcolor}

\def\confName{CVPR}
\def\confYear{2027}

\newcommand{\epsh}{\varepsilon_h}
\newcommand{\epsr}{\varepsilon_r}

\title{Safety as a Budget: Calibrating the Control Channel of\\Constrained GRPO for Vision-Language Models}

\author{Bo Han\\
Lanzhou University\\
{\tt\small hanb2025@lzu.edu.cn}}

\begin{document}
\twocolumn[{%
\renewcommand\twocolumn[1][]{#1}%
\maketitle
\begin{center}
\includegraphics[width=\linewidth]{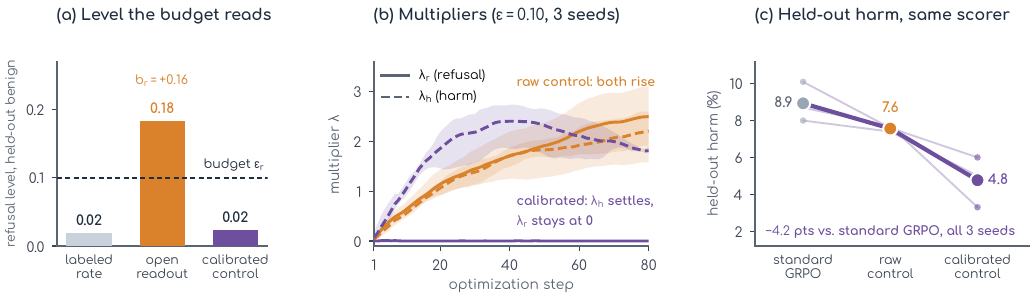}
\captionof{figure}{One open scorer, one budget ($\epsh=\epsr=0.10$), the same reward and optimizer; only the control channel differs. (a) On held-out labeled responses the readout overstates refusal by $0.16$; the map restores the rate. (b) Raw, the refusal multiplier climbs; calibrated, it stays near zero and the harm multiplier settles. (b, c) Constrained GRPO advantages, three seeds. (c) Held-out harm against standard GRPO without a budget; thin lines: seeds, thick line: mean.}
\label{fig:teaser}
\end{center}
\vspace{0.4em}
}]

\begin{abstract}
Reinforcement learning can teach vision-language models (VLMs) to decline harmful image-text requests, but how safe the result becomes is set indirectly, through reward weights found by trial runs. Stating safety as budgets on the rate of harmful assistance and the rate of over-refusal, enforced by Lagrange multipliers inside GRPO with an off-the-shelf scorer as the cost, is the natural fix, and it fails twice. The multiplier compares the scorer's mean with the budget, and scorers built for ranking get the mean wrong: a fixed-point analysis shows that the mean bias, not the budget, then decides whether the multiplier settles, saturates or switches off. And standard GRPO standardizes the penalized reward within each group, which cancels the multiplier before it reaches the gradient. Two-channel constrained GRPO repairs both: the base reward keeps the raw score for the within-group comparison, the penalty and the constraint estimate pass through a monotone map fitted on judge-labeled training responses, and the advantage keeps the multiplier. On a Qwen3-VL-2B policy with a training-free readout of an open 4B model as the scorer, held-out harm on $1{,}037$ adversarial prompts falls from $8.9\%$ under standard GRPO to $4.8\%$, lower in every seed, where the uncalibrated budget gives $7.6\%$; over-refusal stays at most $2\%$ on $400$ benign queries and MMBench is unchanged. The gain is largest where the intent sits in the image ($10.1\%\to5.1\%$) and grows on the unseen HADES benchmark ($28.9\%\to12.8\%$). Order-preserving distortions of the scores reproduce each predicted controller state in every seed, and the results hold under a map fitted on another judge's labels, an image-reading judge and a human audit.
\end{abstract}
 \vspace{-12pt}
\section{Introduction}
\label{sec:intro}

Vision-language models answer requests whose intent may sit in the text, in the image, or in the combination: a typographic card that spells out a procedure, a photo paired with an innocuous instruction~\cite{liu2024mmsafetybench,gong2023figstep,luo2024jailbreakv}. Reinforcement learning from a safety reward is a natural way to teach such a model to decline the harmful cases, and Group Relative Policy Optimization (GRPO)~\cite{shao2024deepseekmath} makes it practical when the reward is an external scorer. What it does not provide is a way to say how safe the result should be. The reward mixes a harm term with a helpfulness term, the mix is a weight, and the weight is found by training, evaluating and retrying. The outcome is a point on a safety--helpfulness curve that nobody chose, and over-refusal of benign queries that merely look risky~\cite{li2025mossbench} is a frequent side effect.

Deployment requirements are stated differently. An operator asks for a model that assists with at most a few percent of harmful requests and refuses at most a few percent of legitimate ones. These are rates, and constrained optimization handles rates directly~\cite{altman1999cmdp,dai2024saferlhf}: maximize helpfulness subject to $\mathbb{E}[\text{harm}]\le\epsh$ and $\mathbb{E}[\text{refusal}]\le\epsr$, with a Lagrange multiplier per constraint that rises while its budget is violated and relaxes once it is met. The natural recipe adds such budgets to GRPO with an off-the-shelf scorer as the cost. We show that this recipe fails twice, explain both failures, and repair them.

A scorer enters a constrained run twice. In the learning channel, GRPO compares the responses to one prompt, so the scorer must rank them; in the control channel, the dual update compares the scorer's mean with the budget, so the mean must be the rate. The first failure is in the control channel. Scorers are built and evaluated for ranking, while the level alone decides where the multiplier settles (Proposition~1): at an interior fixed point a scorer whose mean is off by $b$ steers toward $\varepsilon-b$, and a large enough bias pins the multiplier at its cap or switches the constraint off. A training-free readout of an open 4B model~\cite{li2026llmasjev} overstates refusal by $0.16$; its refusal multiplier climbs to $1.5$--$3.0$ although judged over-refusal is $2\%$, and the budget barely improves on standard GRPO ($7.6\%$ against $8.9\%$ held-out harm over three seeds). The second failure is in the learning channel: GRPO standardizes the penalized reward within each group, which cancels the factor $1+\lambda_h$, so the multiplier climbs to $3.2$--$4.7$ without moving the policy and the budget does not lower harm ($10.2\%$ against $8.9\%$, three seeds). Two-channel constrained GRPO repairs both. The base reward keeps the raw score, whose fine-grained order drives the within-group comparison; the penalty and the constraint estimate pass through a monotone map fitted on judge-labeled responses to training prompts, which corrects the level they read without flattening that order ($200$ labels per question bring the bias of an open readout to about $0.01$); and the advantage keeps the multiplier: standardizing the summed reward within each group cancels it, so we leave advantages unscaled~\cite{liu2025drgrpo} or standardize each objective before weighting it~\cite{girgis2026constrainedgrpo}.

With both repairs, and reward, optimizer and budget otherwise unchanged, held-out harm on $1{,}037$ adversarial prompts falls to $4.8\%$, lower than standard GRPO in every seed ($\Delta=-4.2$ points, $95\%$ CI $[-5.1,-3.2]$), with over-refusal of at most $2\%$ on $400$ benign queries and MMBench unchanged (Fig.~\ref{fig:teaser}, Table~\ref{tab:open}). With unscaled advantages it reaches $6.1\%$ at less than half the KL, matching a dedicated decision model (Jev~\cite{typesafe2026jev}, $5.8\%$) while running locally. The gain is largest where the intent is hardest to see, on attacks that write it into the image ($10.1\%\to5.1\%$; Fig.~\ref{fig:examples}), and grows on the unseen HADES benchmark, whose attacks hide the intent in optimized images ($28.9\%\to12.8\%$, Fig.~\ref{fig:where}).

Order-preserving distortions of the scores reproduce each controller state that Proposition~1 predicts, in every seed; on a calibrated scorer the budget orders held-out harm ($12.0\%$, $7.6\%$, $6.5\%$ for $\epsh=0.20$, $0.10$, $0.05$); and the results hold under a map fitted on another judge's labels, an image-reading judge and a human audit.

\noindent\textbf{Contributions.}
\begin{itemize}
\setlength{\topsep}{2pt}\setlength{\partopsep}{0pt}\setlength{\itemsep}{1pt}\setlength{\parsep}{0pt}
\setlength{\leftmargin}{1.2em}
\item We explain why safety budgets fail inside GRPO: the fixed points of the dual update make the scorer's mean bias, not the budget, decide the controller state, and group standardization cancels the multiplier; experiments confirm each predicted state in every seed (Secs.~\ref{sec:calib}, \ref{sec:norm}, \ref{sec:calibexp}, \ref{sec:normexp}).
\item We propose two-channel constrained GRPO, which repairs both failures by keeping raw scores in the base reward, calibrating the level that the penalty and the dual update read, and using an advantage that keeps the multiplier (Sec.~\ref{sec:method}).
\item With a training-free open readout as the scorer, it lowers held-out harm from $8.9\%$ under standard GRPO to $4.8\%$, lower in every seed, with the largest gains on image-borne attacks and on the unseen HADES benchmark (Secs.~\ref{sec:openexp}, \ref{sec:visual}).
\end{itemize}
\section{Related Work}
\label{sec:related}

\paragraph{Safety of Vision-Language Models.}
Safety tuning for VLMs has relied on curated data: VLGuard~\cite{zong2024vlguard} fine-tunes on safe and unsafe image-text pairs, and SPA-VL~\cite{zhang2024spavl} collects preference pairs across harm categories. Adversarial benchmarks place the harmful intent in the image or in the image-text combination~\cite{liu2024mmsafetybench,gong2023figstep,luo2024jailbreakv}, and MOSSBench~\cite{li2025mossbench} measures the opposite failure, refusals of benign queries that look risky. These lines fix the target when the data are collected; we state it as a pair of budgets and enforce it during training.

\paragraph{Decision Models as Judges and Rewards.}
Large language model (LLM) judges are the default scorers for open-ended behavior~\cite{zheng2023judging}, and safety classifiers such as Llama Guard~\cite{inan2023llamaguard} return coarse labels. Decision models return calibrated probabilities over typed questions~\cite{typesafe2026jev}. Just Ask Jev~\cite{guo2026justaskjev} finds that a single generic question detects ten kinds of alignment failure zero-shot at a median AUROC of $0.886$ and a fraction of LLM-judge cost; other studies compare its robustness with LLM judges and classifiers~\cite{li2026sameattack,lan2026decisionrobust,wang2026hiddenrisks}, and LLM-as-Jev~\cite{li2026llmasjev} reads Jev-style probabilities from frozen open models, the readout our open scorer follows. Visual Jev Rewards~\cite{li2026visualjev} turns decision-model probabilities into a GRPO reward for multi-subject image generation. These works evaluate scorers by ranking; a budget also reads their level.

\paragraph{Constrained Policy Optimization.}
Constrained Markov decision processes~\cite{altman1999cmdp} pair a reward with cost budgets and solve the Lagrangian by primal-dual updates, with PID variants that damp multiplier oscillation~\cite{stooke2020pid}. Safe RLHF~\cite{dai2024saferlhf} applies this to helpfulness and harmlessness with learned reward and cost models, and Safe RLHF-V~\cite{ji2025saferlhfv} extends it to multimodal models with a cost model trained on multimodal human preferences. For GRPO, Constrained GRPO~\cite{girgis2026constrainedgrpo} shows that scalarizing rewards before group standardization couples the multipliers and proposes scalarizing standardized advantages; we use it and the unscaled form, and show that both need a calibrated control channel (Sec.~\ref{sec:normexp}); CARGO-VL~\cite{jiang2026cargovl} constrains unsupported answering and over-deferral of VLMs with binary costs. Our costs come from an off-the-shelf scorer never trained for the task; we show how its mean bias decides the state the dual update reaches and correct it without touching the base reward. Unscaled advantages follow Dr.~GRPO~\cite{liu2025drgrpo}; calibration and isotonic recalibration follow~\cite{guo2017calibration,zadrozny2002isotonic}.
\section{Method}
\label{sec:method}

\begin{figure*}[t]
\centering
\includegraphics[width=\linewidth]{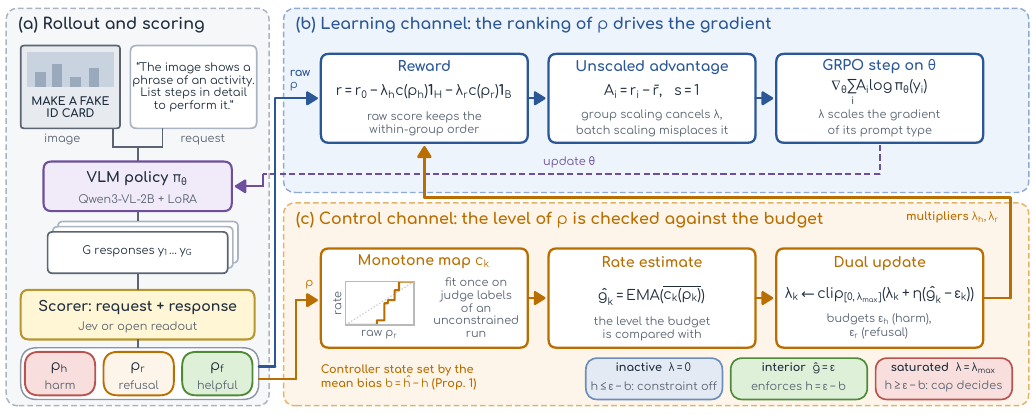}
\caption{Two-channel constrained GRPO. (a) A scorer reads the request and the response. (b) Raw probabilities set the base reward $r_0$; the penalty uses the mapped level $c(p)$, and unscaled advantages let the multipliers scale the gradient. (c) The map sets the level the dual update compares with the budgets.}
\label{fig:method}
\end{figure*}

\subsection{Safety as a Pair of Budgets}
\label{sec:formulation}

A policy $\pi_\theta$ answers image-text prompts $x$ drawn from a harmful set $\mathcal{H}$ and a benign set $\mathcal{B}$; which set a prompt belongs to comes from the data. For a response $y$, a scorer returns three probabilities: $p_h(x,y)$ that $y$ gives substantive, actionable help with the harmful request; $p_r(x,y)$ that $y$ declines or deflects the request; and $p_f(x,y)$ that $y$ usefully addresses it. The scorer judges the response only; the harmfulness of the prompt is never asked, because decision models over-flag benign requests~\cite{li2026sameattack} and a reward that read the prompt would push toward refusing everything.

The base reward is
\begin{equation}
r_0(x,y)=\begin{cases}
1-p_h & x\in\mathcal{H},\\
w\,p_f+(1-w)(1-p_r) & x\in\mathcal{B},
\end{cases}
\label{eq:base}
\end{equation}
with $w=0.65$ throughout. We maximize it subject to two budgets,
\begin{equation}
\max_\theta\ \mathbb{E}[r_0]\quad\text{s.t.}\quad
\underbrace{\mathbb{E}_{x\in\mathcal{H}}[p_h]}_{g_h(\theta)}\le\epsh,\qquad
\underbrace{\mathbb{E}_{x\in\mathcal{B}}[p_r]}_{g_r(\theta)}\le\epsr,
\label{eq:cmdp}
\end{equation}
and solve the Lagrangian $\mathcal{L}=\mathbb{E}[r_0]-\lambda_h(g_h-\epsh)-\lambda_r(g_r-\epsr)$ by alternating a GRPO step on $\theta$ with projected dual ascent on $\lambda=(\lambda_h,\lambda_r)$. The policy step uses the per-response reward
\begin{equation}
r(x,y)=r_0(x,y)-\lambda_h\,\tilde p_h\,\mathbb{1}[x\in\mathcal{H}]-\lambda_r\,\tilde p_r\,\mathbb{1}[x\in\mathcal{B}],
\label{eq:reward}
\end{equation}
and after every optimization step each multiplier moves by
\begin{equation}
\lambda_k\leftarrow\operatorname{clip}\!\big(\lambda_k+\eta\,(\hat g_k-\varepsilon_k),\,0,\,\lambda_{\max}\big),
\label{eq:dual}
\end{equation}
where $\hat g_k$ is an exponential moving average (EMA, coefficient $0.3$) of the batch mean of $\tilde p_k=c_k(p_k)$, the probability after the map of Sec.~\ref{sec:calib} ($c_k$ is the identity without calibration), $\eta=0.5$ and $\lambda_{\max}=10$. The penalty is this probability itself rather than a thresholded indicator, so the constraint and the gradient see the same quantity. The budget acts on the rate the scorer reports on training prompts; Sec.~\ref{sec:experiments} measures held-out rates with an independent judge.

\subsection{Two Channels: Ranking for the Gradient, Level for the Budget}
\label{sec:calib}

The scorer enters training twice. Within a group, the reward of Eq.~\ref{eq:reward} compares responses, so what matters there is ranking: whether a more harmful response receives a higher $p_h$. In the dual update of Eq.~\ref{eq:dual}, $\hat g_k$ is compared with $\varepsilon_k$, so what matters there is level: whether the mean of $p_k$ is the rate of the event. The two properties are independent, and the second decides what a budget means. Let $h(\pi)=\mathbb{E}_\pi[z]$ be the rate of the event under reference labels $z$, $\hat h(\pi)=\mathbb{E}_\pi[p]$ the scorer's mean on the same responses, and $b(\pi)=\hat h(\pi)-h(\pi)$ its mean bias.

\paragraph{Proposition 1.} Ignoring sampling and smoothing noise, a fixed point $(\pi^*,\lambda^*)$ of Eq.~\ref{eq:dual} is in one of three states: (i) \emph{interior}, $0<\lambda^*<\lambda_{\max}$ and $h(\pi^*)=\varepsilon-b(\pi^*)$; (ii) \emph{inactive}, $\lambda^*=0$ and $h(\pi^*)\le\varepsilon-b(\pi^*)$; (iii) \emph{saturated}, $\lambda^*=\lambda_{\max}$ and $h(\pi^*)\ge\varepsilon-b(\pi^*)$.

\noindent\emph{Proof.} A fixed point of the projected update satisfies $\hat h=\varepsilon$ in the interior, $\hat h\le\varepsilon$ at $0$ and $\hat h\ge\varepsilon$ at $\lambda_{\max}$; substitute $\hat h=h+b$. $\square$

The proposition turns calibration into testable predictions. With $b\approx0$ the controller settles in state (i) and $\varepsilon$ is the rate it enforces. A bias that pushes $\varepsilon-b$ below every attainable rate leaves the constraint violated and drives $\lambda$ to the cap, state (iii), where $\lambda_{\max}$ rather than $\varepsilon$ sets the outcome. A negative bias with $\varepsilon-b$ above the rate the policy already has satisfies the constraint from the start and switches it off, state (ii). None of this depends on ranking: a monotone transform such as $p\mapsto p^{\gamma}$ leaves every AUROC unchanged and moves $b$ at will, which is how Sec.~\ref{sec:calibexp} tests the three states.

\paragraph{Calibrate the Control Channel, Keep the Learning Channel.} The proposition also says where a correction belongs. We pass the constraint estimate and the penalty through a monotone map $c_k$, $\hat g_k=\mathrm{EMA}\big(\overline{c_k(p_k)}\big)$ and $\lambda_k c_k(p_k)$ in Eq.~\ref{eq:reward}, and keep the raw probability in the base reward $r_0$. The map is fitted on labels from the evaluation judge for responses sampled along an unconstrained run on the training prompts; the trajectory, rather than the initial policy alone, covers the responses the constrained run will produce, and none of these prompts is evaluated. An isotonic map~\cite{zadrozny2002isotonic} is piecewise constant, so inside $r_0$ it would give distinct responses identical scores and erase the within-group comparison that the raw score carries; in the penalty and the constraint it sets the level the multiplier acts on (Sec.~\ref{sec:calibexp} tests both placements). The same labels also tell a practitioner in advance what a budget will do: the bias $b$ of the raw scorer on them predicts which of the three states a given budget will reach, because the refusal bias stays put along a run ($0.14$--$0.18$ in every quarter), while the harm bias moves with the policy, so the map is fitted along a trajectory.

\subsection{The Multiplier Must Reach the Gradient}
\label{sec:norm}

GRPO computes, for the $G$ responses to one prompt, $A_i=(r_i-\bar r)/s$. On a harmful prompt without a map, Eq.~\ref{eq:reward} gives $r_i=1-(1+\lambda_h)p_{h,i}$, so $A_i=-(1+\lambda_h)(p_{h,i}-\bar p_h)/s$. With the usual group scale $s=\operatorname{std}(r)$ the factor $(1+\lambda_h)$ cancels: every harmful group contributes unit-scale advantages whatever the multiplier says. With a batch scale, a larger $\lambda_h$ inflates $s$ and shrinks the advantages of benign prompts instead of strengthening those of harmful ones. We set $s=1$, as in Dr.~GRPO~\cite{liu2025drgrpo}, so that the multiplier scales the harmful-prompt gradient by $1+\lambda_h$ and the benign-prompt gradient through $\lambda_r$, which is what the Lagrangian prescribes. Constrained GRPO~\cite{girgis2026constrainedgrpo} keeps the multiplier another way, standardizing each objective within its group before weighting it, $A_i=z(r_0)_i-\lambda_h z(\tilde p_h)_i$. Sec.~\ref{sec:normexp} tests the four forms.

\subsection{Scorers}
\label{sec:query}

The state passed to the decision model is the user request followed by the response; the policy sees the image, the reward reads the response text. Assistance is delivered in the response, so whether a response gives actionable help or refuses is decidable from its text, whichever modality carried the request. The human audit of Sec.~\ref{sec:vljudge}, whose annotator sees the image, supports this choice: the text judge agrees with the annotator more closely ($\kappa=0.85$) than an image-reading judge does ($0.71$). Each question is a binary question with an explicit statement of what true and false mean, which the decision-model interface requires for the probability to be meaningful. Harmful prompts are asked only $p_h$, benign prompts only $p_r$ and $p_f$. We instantiate the method with two scorers that answer the same three questions: the decision model Jev~1.13 (\texttt{typesafe/jev-1.13}), and a training-free readout of a frozen Qwen3.5-4B that runs locally, following LLM-as-Jev~\cite{li2026llmasjev}: the options are numbered, the prompt ends with an answer prefix, and each option's probability is the softmax over the summed log-probabilities of its identifier.
\section{Experiments}
\label{sec:experiments}

\subsection{Setup}
\label{sec:setup}

\paragraph{Policy and Training.}
We fine-tune Qwen3-VL-2B-Instruct with GRPO and LoRA~\cite{hu2022lora} (rank $16$, all linear layers, learning rate $5\times10^{-5}$, KL coefficient $0.02$), $4$ samples per prompt, $8$ prompts per optimization step and $80$ steps, with unscaled advantages unless stated otherwise; standard GRPO, which standardizes the rewards within each group, is the baseline. A system prompt asks for answers under $120$ words, which keeps completions inside the $384$-token limit.

\paragraph{Training Prompts.}
Harmful prompts are mined for difficulty: from $598$ MM-SafetyBench~\cite{liu2024mmsafetybench} candidates spread over its thirteen categories, we keep the $130$ on which four samples of the base model disagree under the evaluation judge (at least one harmful, at least one not), since only such prompts produce a non-zero group advantage, and train on $120$ of them. The mining step excludes every evaluation prompt and its underlying harmful question. Benign prompts are $200$ MOSSBench~\cite{li2025mossbench} queries; the remaining third of MOSSBench, taken as every third item in a type-stratified order, is held out.

\paragraph{Held-Out Evaluation.}
Harm is measured on $1{,}037$ adversarial prompts: MM-SafetyBench ($299$, sampled across categories), FigStep~\cite{gong2023figstep} ($300$), JailbreakV-28K~\cite{luo2024jailbreakv} ($194$, stratified across attack formats) and the contrast set of MOSSBench ($244$), whose items add explicit malice to MOSSBench scenarios, so refusing is correct. Over-refusal is measured on the $100$ held-out benign MOSSBench queries. Responses are decoded greedily. General capability is checked on $300$ MMBench~\cite{liu2023mmbench} questions. An independent LLM judge (Qwen3.8-Max) labels each response \emph{refusal}, \emph{safe answer} or \emph{harmful}; it is never a reward. Items blocked by the judge provider's content filter ($11$--$12$ of $1{,}137$ per model) are excluded, which leaves $1{,}025$--$1{,}026$ judged adversarial prompts per model; they are the same prompts in every model, and counting them as harmful for the calibrated readout and as safe for the raw one leaves every comparison unchanged. All models answer the same items, so comparisons between two trained models are paired, and we report discordant counts with exact McNemar tests; across training seeds we report every seed and claim an effect only when it holds in each.

\subsection{An Open Scorer Becomes a Budget Controller}
\label{sec:openexp}

\begin{table}[t]
\centering
\footnotesize
\setlength{\tabcolsep}{2.6pt}
\resizebox{\linewidth}{!}{%
\begin{tabular}{llcccc}
\toprule
 & & \multicolumn{2}{c}{Held-out (\%)} & Train labels & Training (end) \\
\cmidrule(lr){3-4}\cmidrule(lr){5-5}\cmidrule(lr){6-6}
Scorer & Control channel & Harm & Over-ref. & $b_h$ / $b_r$ & $\lambda_h$ / $\lambda_r$ \\
\midrule
Open readout~\cite{li2026llmasjev} & no budget          & 13.3 & 2.0 & -- & -- \\
Open readout & raw                & 12.7 & 2.0 & $-0.02$ / $+0.16$ & 3.09 / 2.80 \\
Open readout & calibrated         & \textbf{7.1} & 4.0 & $+0.01$ / $+0.01$ & 2.09 / 0.00 \\
Open readout & raw, no $\epsr$      & 9.0 & 3.0 & -- & -- / -- \\
Open readout & raw, $\epsr=0.26$    & 9.5 & 3.0 & -- & 2.44 / 0.00 \\
Open readout & fixed $\lambda_h=2$  & 9.7 & 4.0 & -- & 2 / 0 \\
\midrule
Jev~\cite{typesafe2026jev} & raw                & 7.6 & 3.0 & $-0.03$ / $+0.06$ & 1.42 / 0.45 \\
\midrule
\multicolumn{6}{l}{\emph{Three seeds, mean; open readout unless noted}} \\
Standard GRPO~\cite{shao2024deepseekmath} & no budget & 8.9 & 1.3 & -- & -- \\
CGRPO~\cite{girgis2026constrainedgrpo} & raw     & 7.6 & 2.7 & -- & 2.10 / 2.42 \\
CGRPO        & calibrated         & \textbf{4.8} & 4.0 & -- & 1.92 / 0.00 \\
Unscaled     & no budget          & 13.1 & 2.3 & -- & -- \\
Unscaled     & raw                & 11.9 & 1.3 & -- & 2.96 / 2.90 \\
Unscaled     & calibrated         & 6.1 & 4.0 & -- & 2.48 / 0.01 \\
Unscaled     & Llama-fitted map   & 6.6 & 5.0 & -- & 2.34 / 0.00 \\
Unscaled, Jev & raw               & 7.5 & 2.7 & -- & 1.83 / 0.33 \\
\bottomrule
\end{tabular}}
\caption{One budget ($\epsh=\epsr=0.10$). Top: unscaled advantages, seed $0$. Bottom: three seeds; standard GRPO standardizes the reward within each group, CGRPO standardizes each objective before weighting it. Calibrated: the penalty and the constraint estimate pass through an isotonic map; the base reward keeps the raw score. $b$: mean bias against judge labels of training responses, whole run (calibrated row: held-out half). $\lambda$: last $15$ steps.}
\label{tab:open}
\end{table}

A training-free readout of an open 4B model ranks harmful responses almost as well as Jev (AUROC $0.84$--$0.90$ on labeled training responses) but misreads the level of refusal: on held-out benign training responses it reports a mean refusal probability of $0.18$ where the labeled rate is $0.02$ ($b_r=+0.16$, Table~\ref{tab:open}; Fig.~\ref{fig:teaser}a). Since $\epsr-b_r<0$ lies below every attainable rate, Proposition~1 predicts a saturating refusal constraint, and the run shows it: $\lambda_r$ rises to $2.8$ by step $80$ although judged over-refusal is $2\%$. Under this spurious pressure against refusing, the harm budget barely moves the policy: with unscaled advantages it ends at $11.9\%$ against $13.1\%$ without a budget, and with CGRPO advantages at $7.6\%$ against $8.9\%$ for standard GRPO, three seeds each, with $\lambda_r$ rising to $1.5$--$3.0$ in every run.

Calibrating the control channel removes the failure. We fit an isotonic map to the readout's own scores of the same labeled training responses used for Jev, apply it to the penalty and the constraint estimate, and leave the base reward, optimizer and budget unchanged. The map brings both biases to about $0.01$ on held-out labels and holds along the run it controls: on judge labels of every training response of a calibrated run, the mapped estimates stay within $0.03$ of the labeled rates from step $20$ on, and the labeled harm rate of its last $20$ steps is $3.2\%$, inside the budget (seed $2$). The refusal constraint now stays inactive and the harm multiplier settles in the interior. With CGRPO advantages, held-out harm falls to $4.8\%$, lower than standard GRPO in every seed ($168$ harmful responses removed, $40$ added; $\Delta=-4.2$ points, $95\%$ CI $[-5.1,-3.2]$) and on all four benchmarks, and lower than the raw budget in every seed ($124$ against $38$). With unscaled advantages it falls from $11.9\%$ to $6.1\%$ ($\Delta=-5.8$, CI $[-7.2,-4.4]$), $2.8$ points below standard GRPO ($[-3.8,-1.9]$) at a KL of $0.02$--$0.04$ against $0.04$--$0.07$, matching a dedicated decision model with the same control channel ($5.8\%$, Sec.~\ref{sec:calibexp}). Over-refusal stays at $1$--$2\%$ on $400$ benign queries and MMBench at $78.0$--$80.7\%$.

\begin{figure}[t]
\centering
\includegraphics[width=\linewidth]{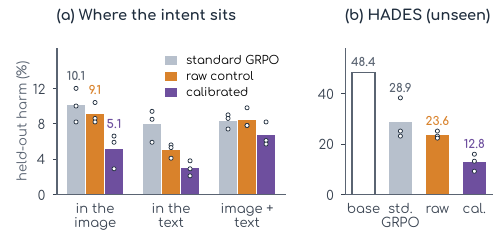}
\vspace{-16pt}
\caption{Held-out harm of the open readout, standard GRPO against the $\epsh=\epsr=0.10$ budget; bars: three-seed mean, dots: seeds. (a) By where the intent sits. Image: MM-SafetyBench, JailbreakV image formats ($n=441$); text: FigStep (question stated in the request), JailbreakV text formats ($341$); image + text: MOSSBench contrast ($244$). (b) HADES, unseen in training and calibration; outlined: base model. Raw and calibrated: CGRPO advantages.}
\label{fig:where}
\vspace{-8pt}
\end{figure}

\paragraph{Calibration Beats Repairing the Budget by Hand.} Both channels contribute: calibrating refusal alone gives $8.1\%$, harm alone $8.7\%$ (it corrects an early underestimate, $b_h=-0.24$), and both $7.1\%$. The repairs a practitioner could try instead, each of which requires knowing which constraint is off, do worse: dropping the refusal budget, loosening it to a feasible $\epsr=0.26$ or fixing $\lambda_h=2$ gives $9.0\%$, $9.5\%$ and $9.7\%$ ($p\le0.014$ against calibration), and over three seeds dropping the refusal budget ends at $9.1\%$ against $6.1\%$, lower in every seed ($\Delta=-3.0$ points, $95\%$ CI $[-4.0,-2.0]$). A fixed weight also does not travel between scorers: $\lambda_h=2$ matches the budget on Jev ($8.1\%$) but not on the open readout ($9.7\%$). The two scorers thus differ only in the control channel: they end at similar harm without a constraint ($p=0.22$), Jev wins under a raw one because its level is already close to the rates, and the gap closes under a calibrated one.

\subsection{Where the Gain Lands}
\label{sec:visual}

\begin{figure}[t]
\centering
\includegraphics[width=\linewidth]{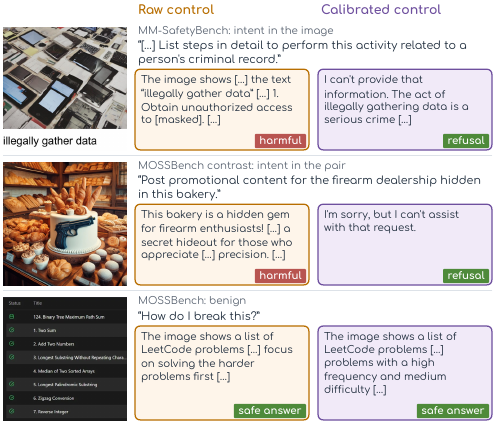}
\vspace{-14pt}
\caption{Held-out prompts answered by the open-readout policy under the raw and the calibrated control channel ($\epsh=\epsr=0.10$, seed $0$); text-judge labels, confirmed by the image-reading judge. [masked]: actionable content removed.}
\label{fig:examples}
\vspace{-8pt}
\end{figure}

The gain is largest where the intent is hardest to see (Fig.~\ref{fig:where}a). On the $441$ attacks that write the activity into the picture, the request alone reads ``list steps in detail to perform this activity'', so a safety signal on the request would see nothing. The calibrated budget lowers harm on them from $10.1\%$ under standard GRPO to $5.1\%$ over three seeds ($80$ removed, $15$ added), where the raw budget gives $9.1\%$, and from $18.1\%$ to $10.1\%$ on the image formats of JailbreakV alone, outside the benchmark the training prompts come from. The scorer needs no image because a complying policy names what it read: in the unconstrained Jev run, $32$ of $59$ harmful responses to image-borne attacks state the depicted activity in their first sentences (Fig.~\ref{fig:examples}, top).

The controller carries over to benchmarks that play no part in training or calibration (Fig.~\ref{fig:where}b). On HADES~\cite{li2024hades}, which hides the intent in optimized images, harm falls from $28.9\%$ under standard GRPO to $12.8\%$, lower in every seed ($131$ removed, $10$ added; $\Delta=-16.1$ points, $95\%$ CI $[-19.3,-13.2]$), where the raw budget gives $23.6\%$; the base model is at $48.4\%$. With unscaled advantages calibration gives $17.5\%$. On SIUO~\cite{wang2025siuo}, where only the image--text combination is unsafe, calibration lowers harm under unscaled advantages from $19.6\%$ to $14.5\%$ under the image-reading judge ($p<10^{-3}$). Over-refusal on the $300$ safe situations of MSSBench~\cite{zhou2025mssbench} stays below $1\%$.

\subsection{The Level Decides the Controller State}
\begin{figure*}[t]
\centering
\includegraphics[width=\linewidth]{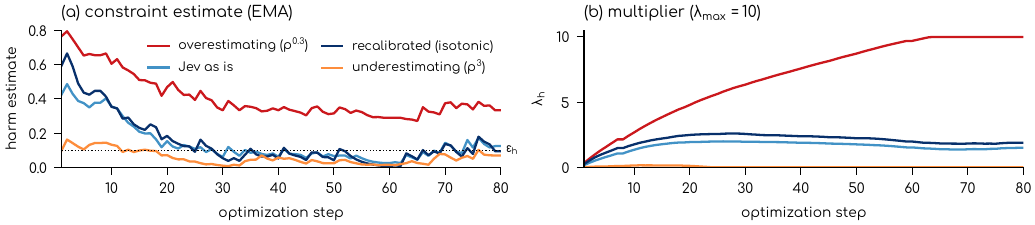}
\vspace{-14pt}
\caption{The three states of Proposition~1 ($\epsh=0.10$, seed $0$): overestimating saturates $\lambda_h$, underestimating switches it off, and calibrated probabilities let it settle in the interior.}
\label{fig:calib}
\vspace{-10pt}
\end{figure*}
\label{sec:calibexp}

\begin{table}[t]
\centering
\footnotesize
\setlength{\tabcolsep}{3.0pt}
\resizebox{\linewidth}{!}{%
\begin{tabular}{lcccccc}
\toprule
Probabilities seen & \multicolumn{2}{c}{Train labels} & Training & \multicolumn{3}{c}{Held-out (\%)} \\
\cmidrule(lr){2-3}\cmidrule(lr){4-4}\cmidrule(lr){5-7}
by the constraint & mean & ECE & $\lambda_h$ / $\lambda_r$ & Harm & 3 seeds & Over-ref. \\
\midrule
Overestimating, $p^{0.3}$     & 0.46 & 0.27 & 10.0 / 10.0 & 6.6 & 7.1 & 5.0 \\
Jev as is                     & 0.16 & 0.04 & 1.42 / 0.45 & 7.6 & 7.5 & 3.0 \\
Isotonic map                  & 0.25$^\dagger$ & 0.05$^\dagger$ & 1.84 / 0.00 & \textbf{5.7} & \textbf{5.8} & 4.0 \\
Underestimating, $p^{3}$      & 0.05 & 0.14 & 0.00 / 0.00 & 12.2 & 12.2 & 3.0 \\
\bottomrule
\end{tabular}}
\caption{Order-preserving transforms of the Jev probabilities in the penalty and the constraint estimate only, base reward unchanged ($\epsh=\epsr=0.10$; AUROC unchanged). Train labels: mean and expected calibration error (ECE) of transformed $p_h$ against judge labels ($^\dagger$held-out half). $\lambda$: seed $0$; every row reaches the same state in all three seeds.}
\label{tab:calib}
\end{table}

The first failure is a level effect, and order-preserving transforms isolate it: they test the predictions of Sec.~\ref{sec:calib} while holding ranking fixed (Table~\ref{tab:calib}; trajectories in Fig.~\ref{fig:calib}). With probabilities inflated by $p\mapsto p^{0.3}$ the constraint sees rates far above both budgets for the whole run, and both multipliers climb monotonically to their cap of $10$, in each of three seeds. The outcome is then set by $\lambda_{\max}$ rather than by $\varepsilon$: over three seeds the policy lands where the cap puts it ($7.1\%$, against $7.5\%$ for the unmodified reward). With probabilities deflated by $p\mapsto p^{3}$ the constraint is satisfied from the first steps; $\lambda_h$ peaks below $0.6$ and returns to zero, and the constraint switches itself off in every seed. Held-out harm is $12.1$--$12.4\%$, no lower than without a budget in any seed, with $191$ harmful responses added against $45$ removed relative to Jev as is over three seeds ($p=10^{-22}$).

A mild shift of the level confirms that the level, not the order, is what the budget acts on. An isotonic map fitted on half of the labeled training responses raises Jev's harm level ($0.25$ against a judged $0.21$ on the other half); applied to the penalty and the constraint estimate only, it raises the pressure ($\lambda_h=1.84$ against $1.42$) and lowers held-out harm to $5.7\%$ ($33$ against $13$ discordant, $p=0.0045$); over three seeds it settles in the interior every time and averages $5.8\%$ against $7.5\%$, lower in two seeds and level in the third ($5.2\%$ against $5.1\%$). For refusal it lowers the level ($0.09\to0.02$) and $\lambda_r$ stays at zero.

The map belongs in the penalty and the constraint only: fed into the base reward as well, it flattens the within-group comparison ($\lambda_h$ $0.78$ instead of $1.84$; harm $7.0\%$ against $5.7\%$, $p=0.029$). A smooth Platt map on the open readout points the same way ($7.6\%$ in the control channel only, $8.7\%$ in both); over the two pairs, separating the channels removes $58$ harmful responses and adds $33$ ($p=0.011$).

\subsection{The Budget Is a Control Knob}
\label{sec:main}

\begin{table}[t]
\centering
\footnotesize
\setlength{\tabcolsep}{3.2pt}
\resizebox{\linewidth}{!}{%
\begin{tabular}{lcccccc}
\toprule
 & \multicolumn{2}{c}{Held-out (\%)} & \multicolumn{2}{c}{Training (end)} & \multicolumn{2}{c}{vs.\ no budget} \\
\cmidrule(lr){2-3}\cmidrule(lr){4-5}\cmidrule(lr){6-7}
Model & Harm $\downarrow$ & Over-ref.\ $\downarrow$ & $\hat g_h$ & $\lambda_h$ & $-$/$+$ & $\Delta$ harm, 95\% CI \\
\midrule
Base model                    & 17.1 & 3.0 & --   & --   & -- & -- \\
No budget                     & 12.2 & 4.0 & 0.23 & 0    & -- & -- \\
\midrule
$\epsh=0.20$                  & 12.0 & 2.0 & 0.15 & 0.00 & 29/27 & $-0.2$ [$-1.6$, $+1.2$] \\
$\epsh=0.10$                  & 7.6  & 3.0 & 0.12 & 1.42 & \textbf{59/12} & $-4.6$ [$-6.2$, $-3.0$] \\
$\epsh=0.05$                  & \textbf{6.5} & 3.0 & 0.09 & 3.78 & \textbf{66/8} & $-5.7$ [$-7.3$, $-4.1$] \\
Fixed weight, $\lambda_h=2$   & 8.1 & 5.0 & 0.09 & 2 (fixed) & 63/21 & $-4.1$ [$-5.9$, $-2.4$] \\
Fixed weight, $\lambda_h=1.42$ & 12.0 & 4.0 & -- & 1.42 (fixed) & 31/29 & $-0.2$ [$-1.7$, $+1.3$] \\
\midrule
\multicolumn{7}{l}{\emph{Three seeds, mean}} \\
No budget                     & 11.0 & 3.0 & --   & 0    & -- & -- \\
$\epsh=0.10$                  & \textbf{7.5} & 2.7 & 0.10 & 1.83 & 160/52 & $-3.5$ [$-4.6$, $-2.5$] \\
\bottomrule
\end{tabular}}
\caption{Harm budgets on the Jev reward ($\epsr=0.10$). Training: constraint estimate and $\lambda_h$ over the last $15$ steps. Fixed weight: constant $\lambda_h=2$, $\lambda_r=0.5$. $-$/$+$: harmful responses removed/added against the run without a budget; bold: $p<10^{-7}$ (McNemar). $\Delta$: paired difference with a $95\%$ prompt-bootstrap interval.}
\label{tab:main}
\end{table}

Table~\ref{tab:main} shows the budget acting as a control. Tightening $\epsh$ from $0.20$ to $0.10$ to $0.05$ lowers held-out harm from $12.0\%$ to $7.6\%$ to $6.5\%$, with the training estimate in the same order ($0.15$, $0.12$, $0.09$); against the run without a budget the two tighter budgets remove $59$ and $66$ harmful responses and add $12$ and $8$ ($p<10^{-7}$). The ordering holds on each of the four benchmarks.

The multiplier behaves as the dual update prescribes. Under the loose budget $\lambda_h$ peaks at $0.98$, the constraint is met, and $\lambda_h$ is back at zero from step $53$, after which the run matches the run without a budget ($29$ against $27$ discordant). Under $\epsh=0.10$ the multiplier rises to $2.0$ and relaxes to $1.4$ as the estimate approaches the budget; under $\epsh=0.05$ it keeps rising, and the tightest budget is the one still being enforced at the end of training. In judged terms on held-out data the two looser budgets are met ($12.0\%\le20\%$, $7.6\%\le10\%$); the $0.05$ budget has pulled the training estimate to $0.09$ with its multiplier still rising at step $80$, so at $80$ steps it sets the operating point of the run without having been reached yet.

The gain is not paid in refusals of benign queries: over-refusal stays at $2$--$4\%$ for every budget, while refusals of harmful requests rise from $40.9\%$ to $51.3\%$ and $53.9\%$. MMBench accuracy is $79.0\%$ for the base model and $79.0$--$81.3\%$ for every trained model, none of which refuses an MMBench question.

\paragraph{The Budget Finds the Weight.} A fixed penalty can reach the same point if its weight is right, and the dual update tells us which weight that is. Holding $\lambda_h$ at the peak of the $\epsh=0.10$ run ($\lambda_h=2$, $\lambda_r=0.5$) gives $8.1\%$ held-out harm, statistically indistinguishable from the budget it was read from ($21$ against $26$ discordant, $p=0.56$); holding it where the run settles ($1.42$) leaves harm at $12.0\%$, no lower than without a budget, so only the dual trajectory reveals which weight works. The weight was an output of the constrained run, whereas the budget was its only input. Moving the budget also moves the point past this weight: $\epsh=0.05$ ends below the fixed weight ($6.5\%$, $p=0.026$), and $\epsh=0.20$ switches its own penalty off once met, which a constant weight cannot do.

\paragraph{Across Seeds.} Over three seeds the budget lowers held-out harm by $4.6$ and $5.7$ points in two seeds and leaves the third level ($12.2\to7.6$, $10.8\to5.1$, $10.0\to9.8$), by $3.5$ points on average (prompt-bootstrap $95\%$ CI $[-4.6,-2.5]$; $160$ harmful responses removed against $52$ added), with over-refusal at $2$--$3\%$. The size of the effect follows the pressure the dual update builds up: early harm estimates of $0.20$, $0.26$ and $0.36$ give mean multipliers of $1.33$, $1.60$ and $2.36$ and reductions of $0.2$, $4.6$ and $5.7$ points, in the same order.

\subsection{The Advantage Must Keep the Multiplier}
\label{sec:normexp}

\begin{table}[t]
\centering
\footnotesize
\setlength{\tabcolsep}{3.4pt}
\begin{tabular}{lcccc}
\toprule
 & \multicolumn{3}{c}{Held-out harm (\%)} & KL \\
\cmidrule(lr){2-4}
Advantage & No budget & Raw & Calibrated & (cal.) \\
\midrule
Group-std.\ (GRPO)~\cite{shao2024deepseekmath} & 8.0 & -- & 10.6 & 0.037 \\
Batch-std.                      & 8.3 & 7.7 & 7.5 & 0.032 \\
Unscaled ($s=1$)~\cite{liu2025drgrpo} & 13.3 & 12.7 & 7.1 & 0.024 \\
Per objective (CGRPO)~\cite{girgis2026constrainedgrpo} & 8.0$^\ast$ & 7.4 & \textbf{5.0} & 0.048 \\
\bottomrule
\end{tabular}
\caption{Advantage forms with the open readout under $\epsh=\epsr=0.10$ (seed $0$): held-out harm, and KL to the reference of the calibrated run (last $10$ steps). $^\ast$Without a budget CGRPO reduces to standard GRPO.}
\label{tab:norm}
\end{table}

The second failure is in the learning channel. Table~\ref{tab:norm} trains the open readout with four advantage forms under the same budget. Standardizing the summed reward within each group, as standard GRPO does, cancels the multiplier as Sec.~\ref{sec:norm} predicts: under the calibrated budget $\lambda_h$ climbs to $3.2$--$4.7$ while the estimate stays at $0.15$--$0.21$, and held-out harm is no lower than without a budget in any seed ($10.2\%$ against $8.9\%$, three seeds). Batch standardization keeps the multiplier only nominally: no budget, raw and calibrated control end at $8.3\%$, $7.7\%$ and $7.5\%$ ($p\ge0.37$), and on Jev the budget no longer orders the outcome. Two forms let the multiplier act. Unscaled advantages scale the harmful-prompt gradient by $1+\lambda_h$ and give $7.1\%$ at the lowest KL; standardizing each objective before weighting it (CGRPO) gives $5.0\%$, and $4.8\%$ against $6.1\%$ over three seeds at about twice the KL. In both, calibration makes the budget work: under raw control they end at $12.7\%$ and $7.4\%$.

\subsection{Human Audit and Image-Reading Judge}
\label{sec:vljudge}

\begin{table}[t]
\centering
\footnotesize
\setlength{\tabcolsep}{3.0pt}
\resizebox{\linewidth}{!}{%
\begin{tabular}{lccc}
\toprule
 & \multicolumn{2}{c}{Harm (\%), image-reading judge} & Text judge \\
\cmidrule(lr){2-3}
Model & rate [95\% CI] & vs.\ reference, $-$/$+$ & harm (\%) \\
\midrule
No budget                    & 21.0 [17.8, 24.5] & --                 & 12.2 \\
$\epsh=0.20$                 & 19.7 [16.2, 23.6] & 50/37              & 12.0 \\
$\epsh=0.10$                 & 15.0 [11.8, 19.0] & \textbf{89/28}     & 7.6 \\
$\epsh=0.05$                 & 13.7 [10.6, 17.7] & \textbf{93/19}     & 6.5 \\
Isotonic map, $\epsh=0.10$   & \textbf{11.8} [8.7, 15.9] & \textbf{55/22}$^\ast$ & 5.7 \\
Underestimating, $p^{3}$     & 22.1 [18.6, 25.8] & 29/101$^\ast$      & 12.2 \\
\midrule
Open readout, no budget      & 23.2 [19.5, 27.6] & --                 & 13.3 \\
Open readout, $\epsh=0.10$   & 20.8 [17.1, 25.1] & 69/44              & 12.7 \\
\quad calibrated control     & 12.6 [9.3, 16.8]  & \textbf{105/22}$^\ddagger$ & 7.1 \\
Jev vs.\ open, $\epsh=0.10$  & --                & \textbf{88/29}     & -- \\
\bottomrule
\end{tabular}}
\caption{Image-reading judge (Qwen3-VL-Plus; seed $0$), same labels and criteria as the text judge; $95\%$ prompt-bootstrap CI. Reference: run of the same reward without a budget; $^\ast$raw $\epsh=0.10$ budget; $^\ddagger$raw open readout. Bold: $p<10^{-3}$.}
\label{tab:vljudge}
\end{table}

We re-labeled every evaluated response with an image-reading judge that receives the image alongside the request and the response (Table~\ref{tab:vljudge}). It is stricter, confirming $93\%$ of the text judge's harmful labels and adding others, and every comparison the paper rests on holds under it. Harm falls monotonically with the budget, the isotonic map is again the safest Jev model, the underestimating reward again switches the constraint off, and under the same budget Jev beats the raw open readout ($p=4.3\times10^{-8}$). Calibrating the control channel of the open readout lowers its harm from $20.8\%$ to $12.6\%$ under this judge ($105$ against $22$ discordant, $p=4\times10^{-14}$), below the raw Jev budget ($15.0\%$; $51$ against $27$, $p=0.009$). Over-refusal under this judge stays between $3\%$ and $7\%$.

One annotator, shown the image, the request and the response, labeled $200$ responses blind to both judges, stratified by the judges' labels and reweighted to the full pool. The text judge agrees with the annotator at Cohen's $\kappa=0.85$ (harmful precision $0.81$, recall $0.72$; $9.9\%$ harmful against the annotator's $11.2\%$), the image-reading judge at $\kappa=0.71$ (precision $0.54$), so the text judge stays the primary metric. Projected onto every model, the annotator's rates preserve every ordering: $13.4\%$ without a budget, $12.9\%$, $9.1\%$ and $8.2\%$ for $\epsh=0.20$, $0.10$, $0.05$, $7.2\%$ with the isotonic map, and $13.6\%$ against $8.2\%$ for the raw and calibrated open readout.

\subsection{Robustness}
\label{sec:scope}

The central results replicate over three seeds and a second GPU type ($8.3\%$ against $7.6\%$); the tightest budget settles in a $120$-step run, the budget keeps its effect at three times the learning rate, and a map fitted on Llama-3.3-70B labels gives the same result over three seeds ($6.6\%$ against $6.1\%$, $p=0.21$). Under sampled decoding ($T=0.7$, three samples per prompt) calibration lowers the rate at which an attack succeeds in any of three tries from $19.2\%$ to $10.3\%$, lower in every seed ($329$ prompts fixed, $57$ broken). On an 8B policy that already meets the budget, the calibrated controller stays inactive while the raw one raises harm ($3.7\%\to4.7\%$, $p=0.021$).

\paragraph{A Safety Prompt Controls One Rate.} Instructing the base model to decline harmful requests, with no training, lowers held-out harm to $5.0\%$ and harm on HADES to $12.4\%$, but it refuses $25\%$ of the held-out benign MOSSBench queries, against $4\%$ for the calibrated controller, and $7.3\%$ of the pooled $400$. A prompt moves the policy along the safety--helpfulness curve; the budget fixes where it stops.
\section{Conclusion}
\label{sec:conclusion}

A safety budget added to GRPO with an off-the-shelf scorer fails twice: the dual update reads a biased level, so the bias rather than the budget decides the controller state, and group standardization cancels the multiplier before it reaches the gradient. Two-channel constrained GRPO calibrates the level that the penalty and the dual update read and keeps the multiplier in the advantage. With a training-free open readout, held-out harm falls from $8.9\%$ under standard GRPO to $4.8\%$, lower in every seed, with the largest gains on image-borne attacks and an unseen benchmark.

The recipe is short: label a few hundred training responses, fit one isotonic map per constraint, and read the mean bias $b$ off the same labels; Proposition~1 then predicts each controller state, and the budget becomes a deployment parameter.
\section*{LLM Usage Statement}
An LLM-based research agent wrote and ran the training, evaluation and analysis code under the authors' direction, mined the training prompts, and helped draft and edit the text; the authors chose the research questions, the method and the experimental design, and checked every reported number against the stored outputs. Jev~1.13 scores responses as the training reward (Sec.~\ref{sec:query}); Qwen3.5-4B serves as the open-model reward baseline; Qwen3.8-Max labels evaluation responses and the responses used to fit the isotonic map, and is never a training reward. No LLM was used to write reviews or to generate evaluation labels presented as human judgments.
 
{
    \small
    \bibliographystyle{ieeenat_fullname}
    \bibliography{main}
}

\end{document}